\documentclass[journal]{IEEEtran}
\usepackage[table]{xcolor}      
\usepackage{booktabs}
\usepackage{graphicx}
\usepackage{makecell}
\usepackage{multirow}
\usepackage{threeparttable}
\usepackage{pgfplots}
\usepackage{color}
\usepackage{xcolor}
\definecolor{rblue}{rgb}{0,0.5,1}
\usepackage{hyperref}
\hypersetup{colorlinks=true,linkcolor=red,citecolor=rblue}

\begin{document}
\title{
FocusFlow: Boosting Key-Points Optical Flow Estimation for Autonomous Driving
}

\author{Zhonghua Yi$^{1,}$\IEEEauthorrefmark{1}, Hao Shi$^{1,4,}$\IEEEauthorrefmark{1}, Kailun Yang$^{2,3,}$\IEEEauthorrefmark{2}, Qi Jiang$^1$, Yaozu Ye$^1$, Ze Wang$^1$, Huajian~Ni$^4$,\\and Kaiwei Wang$^{1,}$\IEEEauthorrefmark{2}%
\thanks{This work was supported in part by the National Natural Science Foundation of China (NSFC) under Grant No. 12174341, in part by Shanghai SupreMind Technology Company Ltd., and in part by Hangzhou SurImage Technology Company Ltd.}
\thanks{$^1$Z. Yi, H. Shi, Q. Jiang, Y. Ye, Z. Wang, and K. Wang are with the State Key Laboratory of Extreme Photonics and Instrumentation, Zhejiang University, Hangzhou 310027, China (email: yizhonghua@zju.edu.cn; haoshi@zju.edu.cn; qijiang@zju.edu.cn; yaozuye@zju.edu.cn; 22030039@zju.edu.cn; wangkaiwei@zju.edu.cn).}%
\thanks{$^2$K. Yang is with the School of Robotics, Hunan University, Changsha 410012, China (email: kailun.yang@hnu.edu.cn).}%
\thanks{$^3$K. Yang is also with the National Engineering Research Center of Robot Visual Perception and Control Technology, Hunan University, Changsha 410082, China.}%
\thanks{$^4$H. Shi and H. Ni are with Shanghai SUPREMIND Technology Company Ltd., Shanghai 201210, China (email: shihao@supremind.com; nihuajian@supremind.com).}%
\thanks{\IEEEauthorrefmark{1}denotes equal contribution.}%
\thanks{\IEEEauthorrefmark{2}corresponding authors: Kaiwei Wang and Kailun Yang.}%
}

\markboth{IEEE Transactions on Intelligent Vehicles, September~2023}%
{Yi \MakeLowercase{\textit{et al.}}: FocusFlow}
\maketitle

\begin{abstract}
Key-point-based scene understanding is fundamental for autonomous driving applications. At the same time, optical flow plays an important role in many vision tasks. However, due to the implicit bias of equal attention on all points, classic data-driven optical flow estimation methods yield less satisfactory performance on key points, limiting their implementations in key-point-critical safety-relevant scenarios. To address these issues, we introduce a points-based modeling method that requires the model to learn key-point-related priors explicitly. Based on the modeling method, we present FocusFlow, a framework consisting of 1) a mix loss function combined with a classic photometric loss function and our proposed Conditional Point Control Loss (CPCL) function for diverse point-wise supervision; 2) a conditioned controlling model which substitutes the conventional feature encoder by our proposed Condition Control Encoder (CCE). CCE incorporates a Frame Feature Encoder (FFE) that extracts features from frames, a Condition Feature Encoder (CFE) that learns to control the feature extraction behavior of FFE from input masks containing information of key points, and fusion modules that transfer the controlling information between FFE and CFE. Our FocusFlow framework shows outstanding performance with up to ${+}44.5\%$ precision improvement on various key points such as ORB, SIFT, and even learning-based SiLK, along with exceptional scalability for most existing data-driven optical flow methods like PWC-Net, RAFT, and FlowFormer. Notably, FocusFlow yields competitive or superior performances rivaling the original models on the whole frame. The source code will be available at \url{https://github.com/ZhonghuaYi/FocusFlow_official}.
\end{abstract}

\begin{IEEEkeywords}
Optical flow estimation, autonomous driving, key points, conditional modeling.
\end{IEEEkeywords}

\IEEEpeerreviewmaketitle

\section{Introduction}

\begin{figure}[!t]
    \centering
    \includegraphics[width=\linewidth]{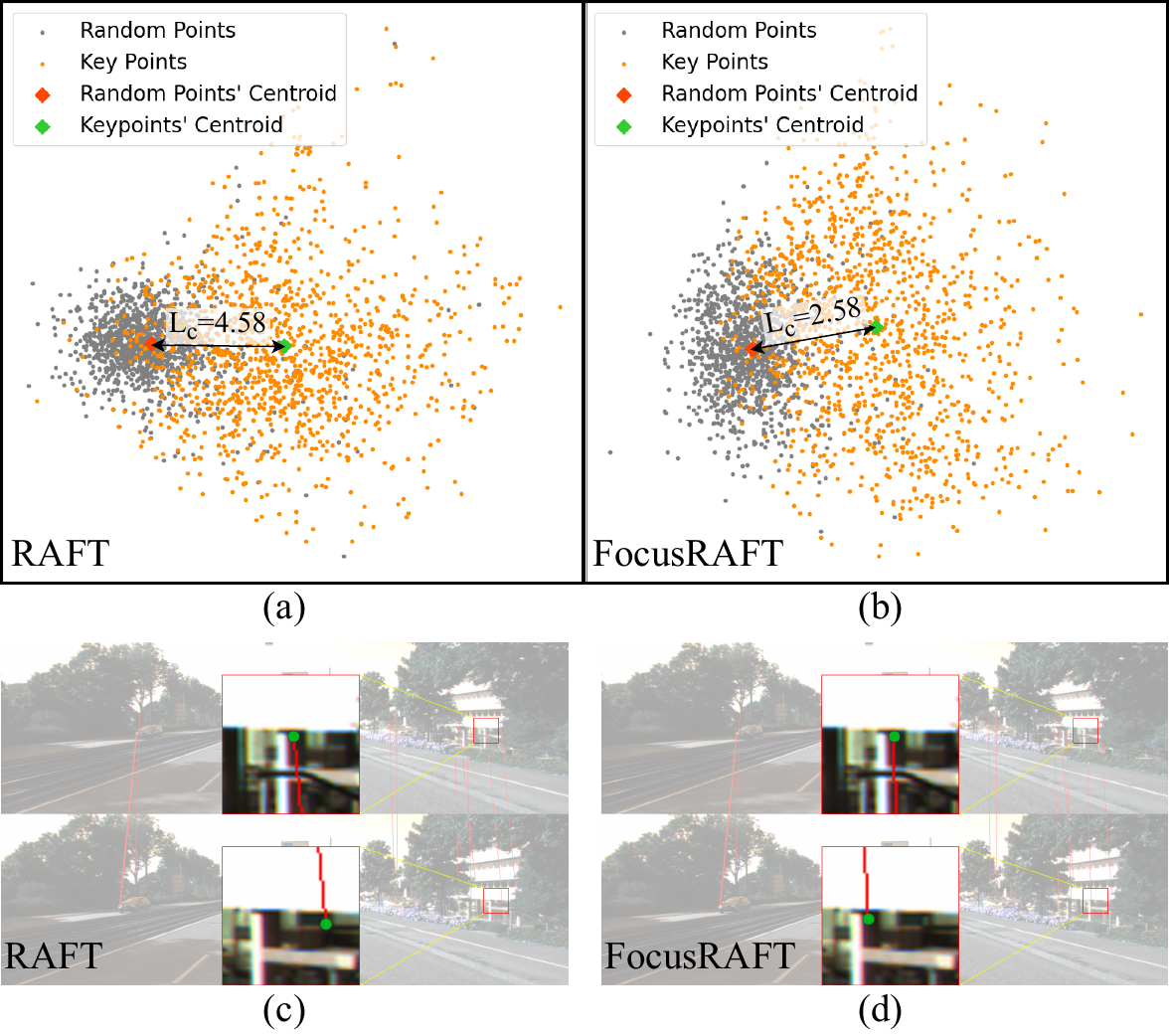}
    \caption{\textbf{Illustration of points feature distribution and effects of the proposed FocusFlow framework.} 
    (a) and (b) illustrate the random and key points' feature distributions where the features of points are from the embedding space of the network's encoder and the dimension is reduced by using PCA~\cite{pearson1901liii}.
    $L_c$ represents the Euclidean distance between the centroids of the two point-feature sets.
    RAFT~\cite{teed2020raft}, a classic representative optical flow network, shows insufficient capacity to encode key points into the same feature space as the random points in (a), which shows that the knowledge learned from the whole frame cannot adapt to the specific key points.
    By integrating RAFT into the FocusFlow framework, a notable reduction in the distance between the centroids of randomly selected points' features and key points' features is observed in FocusRAFT.
    (c) and (d) depict a comparison under an autonomous driving scenario from KITTI~\cite{geiger2013vision}, proving our framework achieves better key point matching results.
    }
    \vspace{-1em}
    \label{fig:pca}
\end{figure}

\IEEEPARstart{O}{ptical} flow estimation is a long-standing problem, which aims to predict per-pixel motion relation between two consecutive frames. 
Providing pixel-level correspondence information, dense optical flow has been used in many navigation-critical tasks, like Simultaneous Localization and Mapping (SLAM)~\cite{caruso2015large,gadde2017semantic,cheng2019improving}, autonomous driving~\cite{ranft2016role,okafuji2016design,shi2022csflow},
object tracking~\cite{dueholm2016trajectories,saleh2018intent,rangesh2019no},
and beyond-field-of-view estimation for scene understanding~\cite{shi2022flowlens}, \textit{etc.}

Recently, owing to the substantial advancements in the field of deep learning, an increasing number of learning-based methodologies have been devised and implemented within the domain of intelligent vehicles. 
For instance, learning-based object detection techniques have found practical application in Unmanned Aerial Vehicles (UAVs)~\cite{liu2022yolov5}.
Additionally, Recurrent Neural Networks (RNNs) have been successfully employed for predicting the intent of pedestrians~\cite{saleh2018intent}.
Hence, it is imperative to conduct research into learning-based optical flow estimation to satisfy the evolving needs of intelligent vehicle systems.

In autonomous driving~\cite{li2020rtm3d,li2018stereo} and SLAM~\cite{davison2007monoslam,mur2015orb}, it is known that key points play an important role and have been widely used.
For example, ORB key points are leveraged in~\cite{mur2015orb} to provide good invariance to changes in viewpoint and illumination, while ordinal key points are predicted in RTM3D~\cite{li2020rtm3d} to represent a 3D object for providing accurate project points for multi-scale objects, which are vital in real-world applications.
On the other hand, key points' optical flow is crucial for upper-level navigation tasks like visual odometry~\cite{qin2018vins}, in which the optical flow of key points is used to estimate the pose of the camera.
This raises a pivotal research question: How to boost optical flow estimation on key points, rendering it more focused and dependable for tasks centered around key points, particularly in the context of autonomous driving?

In recent years, many data-driven optical flow estimation methods~\cite{teed2020raft,sun2018pwc,shi2023panoflow,huang2022flowformer} have been successfully developed, by using Convolutional Neural Networks (CNNs) and vision-transformer-like architectures. 
These works are established under a general idea, including a photometric objective and an encoder-decoder network architecture.
This photometric objective does not perform well when considering the estimation precision at key points, since their optimization goal is to minimize the photometric error on the whole frame, disregarding the distinct distribution of key points.
Besides, the network does not have a key-point-oriented design, which makes it difficult to learn a point-related representation explicitly and correctly.
With the objective of enhancing the utility and efficiency of optical flow estimation for practical applications in autonomous driving, this work is presented.

To explore the best form of estimating the motion of key points, we rethink existing optical flow estimation methods by considering points in the scene as independent samples and put forward a form of joint distribution that indicates a prior related to key points should be learned.
Moreover, to help the model learn the prior from the complicated context, an input mask including information about the key points is utilized.

With the aim of adapting this modeling method in which different types of points require different processing procedures, a Conditional Point Control Loss (CPCL) function is proposed, which produces different attention among given points, and it is combined with an ordinary photometric loss function to learn robust point representations among key points and other points. 
This mix objective is able to significantly improve the upper limit of the model in estimating the optical flow of key points, with a competitive precision on the whole frame, even surpassing the original model in some cases.
Besides, with the change of the proportion between the photometric loss function and CPCL or the supervising area of CPCL, the optimization direction can be easily adjusted to learning estimation on the whole frame or specific types of points.

Then, a controlling model is presented, which explicitly learns conditional control from the above diverse point-wise loss function.
We propose the Conditional Control Encoder (CCE) which contains a conventional Frame Feature Encoder (FFE), a Condition Feature Encoder (CFE), and cross-stream fusion modules, to substitute the classic feature encoder of optical flow networks. 
FFE learns to extract point features from the frame, whereas CFE learns to control the extracting behavior of FFE, by using a bidirectional fusion module at each stage of FFE for enhanced control.
In particular, simple and effective $1{\times}1$ convolutions are implemented as the fusion method.
For the remaining structures, we maintain the architectures in their original forms. 
Hence, this conditioned model is adaptable to a wide range of existing optical flow estimation networks while retaining the inherent attributes of the original design.

Finally, the mixed objective and controlling model are combined into the proposed framework FocusFlow, which learns key points control with CCE under the unequal supervision of points.
We equip it on three representative optical flow networks of different generations,
including PWC-Net~\cite{sun2018pwc} which first uses cost volumes,
RAFT~\cite{teed2020raft} which first uses GRU updaters~\cite{cho2014properties}, and FlowFormer~\cite{huang2022flowformer} which first uses full transformer-based architectures,
and consistently achieve significant precision improvements on various key points and various scenes. 

As shown in Fig.~\ref{fig:pca}(a)(b), the FocusFlow framework achieves a smaller Euclidean distance $L_c$ between the feature centroids of random points and key points.
While conventional optical flow estimation methods focus on the overall points~(\textit{i.e.}, the random points), the FocusFlow framework can focus on the key and overall points at the same time, therefore $L_c$ becomes smaller.
As a result, the FocusFlow framework exhibits superior performance in optical flow estimation for key points, as depicted in Fig.~\ref{fig:pca}(c)(d) and more detailed qualitative comparisons in Fig.~\ref{fig:qualitative}.

Our proposed FocusFlow models outstrip the original models, where the FocusRAFT model has the best estimation precision on random-split Sintel-val dataset~\cite{butler2012naturalistic} and the KITTI-val dataset~\cite{geiger2013vision}, showing greatly enhanced performance on key points and competitive or superior performance on the whole frame rivaling with the original models.
Besides, FocusFlow is able to seamlessly adapt to arbitrary types of key points, including ORB~\cite{rublee2011orb}, SIFT~\cite{lowe2004distinctive}, and even learning-based SiLK~\cite{gleize2023silk}.
Among those four types of key points, the precision improvement reaches as high as ${+}44.5\%$ on the FlyingChairs dataset~\cite{dosovitskiy2015flownet}, highlighting the efficacy and versatility of the proposed FocusFlow framework.

According to comprehensive ablation studies, the sparse binary mask which marks the location of key points is identified as the simplest and most effective option for the input mask.
In addition, compared with the conventional photometric loss, the mix objective achieves ${+}12.99\%$ in precision improvement on key points without any alteration to the model's architecture.
Furthermore, $1{\times}1$ convolutions are proved as the simplest and outperform conventional fusion methods.

At a glance, this work delivers the following contributions:
\begin{enumerate}
    \item A point-based modeling method with explicit consideration of key points' prior probability distribution is introduced.
    \item The Conditional Point Control Loss (CPCL) function and a mixed loss function are proposed, as compared against classic photometric loss functions. Besides, the CPCL can focus on precision improvement at key points in optical flow estimation.
    \item A novel conditional controlling architecture is presented, which is compatible with existing encoder-decoder optical flow estimating architectures.
    \item The FocusFlow, a novel framework that can adapt to the new objective function by using conditional modeling is proposed, which is scalable for the most of existing data-driven optical flow estimation methods and key points.
\end{enumerate}

\section{Related Work}
\begin{figure}[t]
    \centering
    \includegraphics[width=1.0\columnwidth]{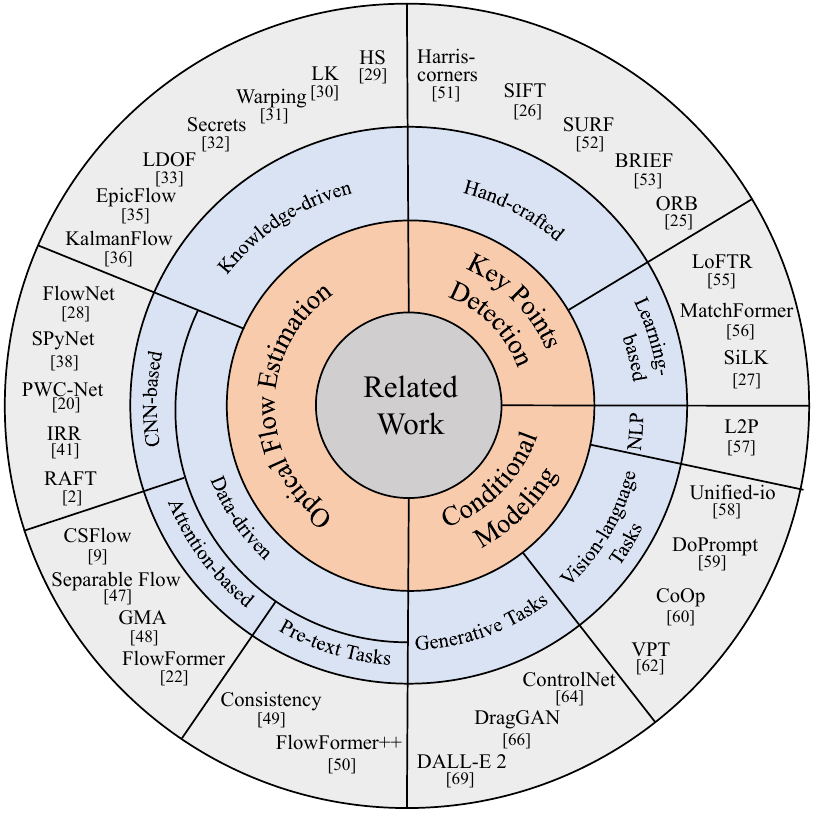}
    \caption{\textbf{Assorted overview of the Related Work section.} Three research areas are mentioned, including optical flow estimation, key points detection, and conditional modeling. Each part has more detailed classifications.}
    \label{fig:related work}
\end{figure}

In this section, we provide a summary and analysis of related works, encompassing three primary areas: optical flow estimation, key points detection, and conditional modeling. We have compiled representative research findings and organized them into an overview figure, as depicted in Fig.~\ref{fig:related work}.

\subsection{Optical Flow Estimation}

\noindent \textbf{Knowledge-driven methods}. 
Horn and Schnuck~\cite{horn1981determining} propose the first optical flow estimation framework, using a variational method to optimize an energy function coupling the brightness constancy and spatial smoothness assumptions and estimate a dense flow field.
Following this strategy, Lucas and Kanade~\cite{lucas1981iterative} introduce the local constraint and estimate a sparse flow field. 
Due to the high-speed moving objects in many scenarios, large displacement and occlusion are serious problems in optical flow estimation.
To deal with large displacement, Brox~\textit{et al.}~\cite{brox2004high} leverage a coarse-to-fine warping strategy to deal with the large displacement problem.
Sun~\textit{et al.}~\cite{sun2010secrets}
introduce a non-local term to classical optical flow objective function, therefore robustly integrating flow estimates over large spatial neighborhoods. Other methods introduce rich descriptors~\cite{brox2010large} or use a hierarchical correspondence field search strategy~\cite{bailer2015flow}.
For the occlusion problem, Revaud~\textit{et al.}~\cite{revaud2015epicflow} propose a sparse-to-dense interpolation scheme, which is robust to motion boundaries, occlusions, and large displacements.
While previous works primarily focus on spatial coherence, Bao~\textit{et al.}~\cite{bao2018kalmanflow,bao2019kalmanflow} propose an efficient estimation method under a Kalman filtering system.
Though these knowledge-driven methods have been developed for years and have resulted in a profound understanding of optical flow estimation, data-driven approaches have shown potential in efficiency and accuracy in recent years.

\noindent \textbf{Data-driven methods}. 
Dosovitskiy~\textit{et al.}~\cite{dosovitskiy2015flownet} are the first to introduce CNNs into optical flow estimation.
They propose FLowNetC and FlowNetS with end-to-end models learned on the synthetic Flying Chairs dataset.
Due to its large model size, the runtime of FlowNet is not satisfactory for mobile applications. 
Ranjan~\textit{et al.}~\cite{ranjan2017optical}, inspired by classical knowledge-based methods, present a spatial-pyramid-based coarse-to-fine approach named SpyNet, by warping the reference image by current estimated flow at each pyramid level and update the flow, which has fewer parameters and higher speed. 
Based on~\cite{ranjan2017optical},
Hu~\textit{et al.}~\cite{hu2018recurrent} and Dai~\textit{et al.}~\cite{dai2018pyramid} present more advanced networks to deal with the significant displacement problem. 
Sun~\textit{et al.}~\cite{sun2018pwc} introduce PWC-Net, which uses cost volume constructed by query features and warped reference features to the pyramid network, leading to a more accurate and faster estimation result. 
Following~\cite{sun2018pwc}, Hur and Roth~\cite{hur2019iterative} propose an Iterative Residual Refinement (IRR) scheme that reduces the number of parameters and improves the accuracy,
Zhai~\textit{et al.}~\cite{zhai2019optical} introduce a self-attention mechanism for better feature representation, while Zhao~\textit{et al.}~\cite{zhao2020maskflownet} use an asymmetric occlusion-aware feature matching module for better flow and occlusion estimation. 
Data-driven methods have been highly developed after the introduction of RAFT~\cite{teed2020raft}, which builds multi-scale 4D correlation volumes for all pairs of pixels and iteratively updates a flow field through a modified GRU block~\cite{cho2014properties} that performs lookups on the correlation volumes. 

Along with the transformers~\cite{vaswani2017attention} in Natural Language Processing (NLP) being successfully adopted in vision tasks~\cite{liu2021swin}, attention-based modules are introduced into optical estimation methods~\cite{zhao2022global} to reach better results, especially when estimating large displacement and occlusion. 
As the simple cost volume faces the disambiguate motion and lack of non-local knowledge,
Zhang~\textit{et al.}~\cite{zhang2021separable} use a separable cost volume module,
Jiang~\textit{et al.}~\cite{jiang2021learning} propose global motion feature aggregation strategy,
and Shi~\textit{et al.}~\cite{shi2022csflow} propose CSFlow, which consists of a Cross Strip Correlation (CSC) module and a Correlation Regression Initialization (CRI) module, to encode global context into correlation volumes and maximally exploit the global context for optical flow initialization. 
Huang~\textit{et al.}~\cite{huang2022flowformer} propose FlowFormer, a transformer-based neural network architecture that tokenizes the 4D cost volume, encodes it into a cost memory, and decodes the cost memory into the estimated optical flow. 

With imposing pre-text tasks showing powerful improvements in various tasks, Jeong~\textit{et al.}~\cite{jeong2022imposing} introduce occlusion consistency, zero forcing, and transformation consistency into optical flow estimation.
Shi~\textit{et al.}~\cite{shi2023flowformer++} adopt the masked cost-volume autoencoding scheme and task-specific masking strategy and reconstruction pre-text task,
which can learn better representations from pre-training.
However, previous data-driven methods fail to fully exploit specific image areas like key points, where rich local information can help the model achieve more accurate performance, with implicit inductive bias that all points of the image have the same motion distribution.
In contrast, our work has taken this into consideration, which leads to a more complete framework for boosting key points' optical flow estimation.

\subsection{Key Points Detection}
Traditional key point detection methods are designed to be robust to viewpoint changes and illumination changes. 
Geometric features, like corners, gradients, and scale-space extrema, are often used in human-designed key points detection and description. 
Harris-corners~\cite{harris1988combined}, SIFT~\cite{lowe2004distinctive}, SURF~\cite{bay2008speeded}, ORB~\cite{rublee2011orb}, or other similar works~\cite{calonder2010brief, shi1994good} are all under these consideration.
As deep neural networks have shown more advanced capacities than hand-engineered representations on various tasks, learning-based methods~\cite{sun2021loftr,wang2022matchformer} are proposed and have shown more stable performance with abundant feature extractions. 
More recently, SiLK~\cite{gleize2023silk}, has shown its simple but competitive performance in diverse settings.
With the wide usage of key points on SLAM systems~\cite{davison2007monoslam,mur2015orb} and autonomous vehicles~\cite{li2018stereo}, our work focuses on boosting optical flow estimating precision on such key points, intending to improve the reliability and practicality of data-driven optical flow estimation methods.

\subsection{Conditional Modeling}
Conditional modeling has been highly regarded by researchers.
It considers adding external conditions to existing and well-trained models, making it more concentrated on downstream tasks. 
Prompt learning, a popular tuning method, has been successfully used in NLP tasks~\cite{wang2022learning} and vision-language tasks~\cite{lu2022unified}, and then in vision tasks~\cite{zheng2022prompt}. 
The key idea of prompt learning is adding a few additional trainable parameters into a frozen pre-trained large model and tuning the whole model on specific tasks. 
Zhou~\textit{et al.}~\cite{zhou2022learning} introduce the usage of dynamic prompts on CLIP~\cite{radford2021learning}, showing promising transferability and stronger domain generalizability. 
Jia~\textit{et al.}~\cite{jia2022visual} present Visual Prompt Tuning (VPT), with only a small amount of trainable parameters in the input space of pure vision models like ViT~\cite{dosovitskiy2020image}, achieving significant performance gains compared to other parameter efficient tuning protocols, and even outperforming full fine-tuning in many cases across model capacities and training data scales.
Recently, more inspiring conditional modeling works have shown their potential in the generative model area, including
ControlNet~\cite{zhangAddingConditionalControl2023} adding conditional terms to diffusion models~\cite{rombach2022high},
DragGAN~\cite{pan2023drag} using motion supervision into StyleGAN2~\cite{karras2020analyzing}, and other successful conditional models~\cite{lu2022conditional,ramesh2022hierarchical}.
Motivated by these works, our proposed FocusFlow framework leverages conditional modeling techniques on existing data-driven optical flow estimation methods with simple but effective control, achieving high generalizability and flexibility.

\section{Method}
In this section, we first review the classic optical flow estimation methods' formal definition in Sec.~\ref{sec:problem definition}.
Then, a new modeling method is put forward in Sec.~\ref{sec:points distribution}.
In order to adapt to this modeling method, we propose a mixed loss function combined with the classic photometric loss function and a novel Conditional Point Control Loss (CPCL) function in Sec.~\ref{sec:loss function}.
After that, a novel network architecture using conditional modeling is presented in Sec.~\ref{sec:conditional modeling}.
Finally, the presented loss function and network architecture are implemented into our proposed framework called FocusFlow (Sec.~\ref{sec:framework}).

\subsection{Problem Definition of Optical Flow Estimation}
\label{sec:problem definition}
Given a pair of images $I_1$, $I_2$, the optical flow denotes the point motion relation between two frames. 
Generally, optical flow $\mathbf{f}\left( \mathbf{x} \right) $ describes the consistency between original point $\mathbf{x}$ in $I_1$ and same point $\mathbf{x}+\mathbf{f}(\mathbf{x})$ in $I_2$ after object moving or change of camera pose, which results in the same point appearing in different location of the camera imaging sensor. 
Optical flow estimation methods are supposed to provide a model $M$ that outputs an optical flow estimation $\mathbf{\hat{f}}$ from a pair of given images $(I_1, I_2)$:
\begin{equation}
\mathbf{\hat{f}}=M\left( I_1,\ I_2 \right) .
\end{equation}

For data-driven methods, the model $M$ is usually learned on a collected dataset, with learnable parameters $\theta$. 
Assume an annotated dataset $(X, Y)$ with $N$ pairs of data, where $X$$=$$\{(I_{11}, I_{12}), (I_{21}, I_{22}), ..., (I_{N1}, I_{N2})\}$ is the set of frame pairs consists of query frame $I_{\_1}$ and reference frame $I_{\_2}$, and $Y$$=$$\{\mathbf{f}_1, \mathbf{f}_2, ..., \mathbf{f}_N \}$ is the set of corresponding optical flow.
The method with a model $M_\theta$ can be defined as:
\begin{equation}
\mathbf{\hat{f}}_n=M_\theta \left( X_n \right),
\end{equation}
where $X_n=(I_{n1}, I_{n2})$ and $n=1, 2, ..., N$. 

Data-driven methods usually use stochastic gradient descent to optimize $\theta$, which calls for a suitable loss function for supervision.
For optical flow estimation, a photometric loss function $\mathcal{L}_p(\mathbf{f}_n, \mathbf{\hat{f}_n})$ is commonly used. 
The form of the photometric loss function $\mathcal{L}_p$ is depicted as follows:
\begin{equation}
\mathcal{L}_p(\mathbf{f}_n, \mathbf{\hat{f}}_n) = \frac{1}{K}\sum_{i=1}^K{||\mathbf{f}_{n,i}-\mathbf{\hat{f}}_{n,i}||_p},
\label{equ: photometric objective function}
\end{equation}
where $i$ is the point index, $K$ is the point number, and $p$ denotes the constant used in calculating $p$-norm, generally being $1$ or $2$.
For coarse-to-fine methods, $\mathcal{L}_p$ at each estimation stage is multiplied by specific weights and summed to form the final loss.

\subsection{Points Distribution}
\label{sec:points distribution}
In this part, the classic frame-based modeling method is expanded into a novel point-based modeling method.
Conventional data-driven optical flow estimation methods treat every single image pair as a sample under the distribution of a scene and estimate the corresponding dense optical flow.
Treating like probabilistic models, these methods generally model joint probability distribution $p(X,Y)$, in which $X$ is the image pair set, and $Y$ is the dense optical flow set.
Unfortunately, this modeling method is frame-based, as each sample of $X$ is the form of two frames, which does not perform well when considering the properties of points in the frame.

Here, with the aim of obtaining the per-point relationship, a new point-based modeling method is proposed.
Instead of considering frames, each point $\mathrm{p}$ in the scene is treated as a sample, and the model is used to estimate the per-point optical flow $\mathbf{f_{\mathrm{p}}}$.
This leads to a new method, which models the joint probability distribution $p(\mathrm{p},\mathbf{f_{\mathrm{p}}})$.

Furthermore, the joint probability distribution can be decomposed as follows:

\begin{equation}
\label{equ:joint distribution}
    p(\mathrm{p},\mathbf{f_{\mathrm{p}}}) = p(\mathrm{p}_k)p(\mathbf{f_{\mathrm{p}}}|\mathrm{p}_k),
\end{equation}

which means it is imperative to learn a prior $p(\mathrm{p}_k)$ describing the distribution over the feature space of points and a likelihood $p(\mathbf{f_{\mathrm{p}}}|\mathrm{p}_k)$ describing the distribution of optical flow when $\mathrm{p}_k$ is conditioned.

Most data-driven optical flow estimation methods~\cite{teed2020raft,sun2018pwc,huang2022flowformer,dosovitskiy2015flownet} have not explicitly modeled this problem, with little consideration on $p(\mathrm{p}_k)$ which degrade as uniform distribution, leaving the model just being an estimation of $p(\mathbf{f_{\mathrm{p}}}|\mathrm{p}_k)$.
What we need to do is explicitly consider learning the prior distribution of key points.
By doing so, a complete inference chain from points to optical flow can be constructed.

Since the contexts of key points are complicated, it is hard to learn the prior blindly, a mask that includes the key points' information is inputted into the model.
To enhance learning accuracy and efficiency, a new learning objective and a new network architecture are presented, as will be described in the next two subsections.

\begin{figure}[t]
    \centering
    \includegraphics[width=1.0\columnwidth]{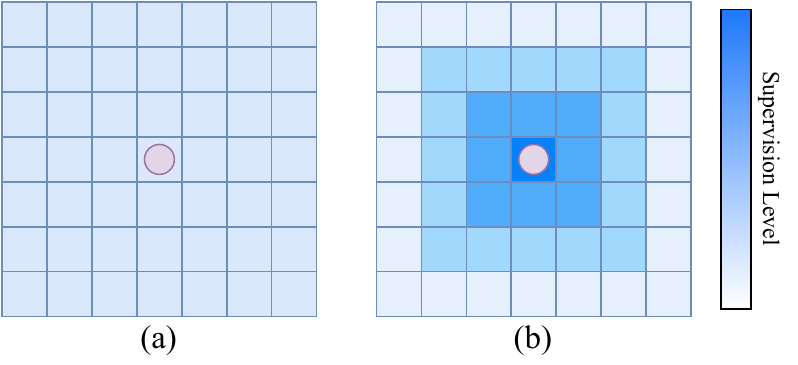}
    \caption{\textbf{Comparison between the classic photometric supervision and the CPCL supervision. The purple dot represents a key point.} (a) Classic photometric supervision, with equal supervision among all points. (b) The proposed CPCL, with diverse supervision levels among points, relates to the Euclidean distance between the current point and key points.}
    \label{fig:supervision}
\end{figure}

\subsection{Loss Function}
\label{sec:loss function}
In this part, we introduce the Conditional Point Control Loss (CPCL) function and combine it with the conventional photometric loss function into a mixed loss function, which we will use as the new optimization objective.

The classic photometric loss function expressed as Eq.~(\ref{equ: photometric objective function}) is with the implicit bias that all points are under the same weight, which is unfavorable to learning $p(\mathrm{p}_k)$ which describes the prior distribution of key points. 
Since we are considering the distribution of key points, it is beneficial to supervise the model with a loss function that provides diverse supervision on different types of points and is related to the interested key points.
Thus, we propose to adapt the form of the normal photometric loss function into a Conditional Point Control Loss (CPCL) function, by using $\alpha_i$ to guide model $M_\theta$ using different attention on different points:

\begin{equation}
\mathcal{L}_{cpcl}(\mathbf{f}_n, \mathbf{\hat{f}}_n) = \frac{1}{\sum_{i=1}^K{\alpha _i}}\sum_{i=1}^K{\alpha _i||\mathbf{f}_{n,i}-\mathbf{\hat{f}}_{n,i}||_p},
\end{equation}

where $\frac{1}{\sum_{i=1}^K{\alpha _i}}$ is a normalization term, which is employed to normalize CPCL into the same value range as $\mathcal{L}_p$.
As shown in Fig~\ref{fig:supervision}, the proposed CPCL effectively supervises the model with diverse point-wise supervision compared with the classic photometric loss function.

The optimization goal of the proposed $\mathcal{L}_{cpcl}$ is not exactly the same as $\mathcal{L}_{p}$.
While $\mathcal{L}_{cpcl}$ focuses on the supervision around the key points, the $\mathcal{L}_{p}$ has better supervision globally which is helpful for dense estimation.
In practice, a mix loss function of $\mathcal{L}_{cpcl}$ and $\mathcal{L}_{p}$ is applied to harness their respective benefits, depicted as follows:

\begin{equation}
\mathcal{L}_{mix}(\mathbf{f}_n, \mathbf{\hat{f}}_n) = \mathcal{L}_p(\mathbf{f}_n, \mathbf{\hat{f}}_n) + \lambda \mathcal{L}_{cpcl}(\mathbf{f}_n, \mathbf{\hat{f}}_n),
\label{equ:mix loss}
\end{equation}

where $\lambda$ is the tuning parameter used to balance the optimization goal between $\mathcal{L}_{cpcl}$ and $\mathcal{L}_{p}$.
The larger $\lambda$ means the optimization goal tends to learn the $\mathcal{L}_{cpcl}$, \textit{i.e.}, the optical flow estimation for key points.

\begin{figure}[t]
    \centering
    \includegraphics[width=0.9\columnwidth]{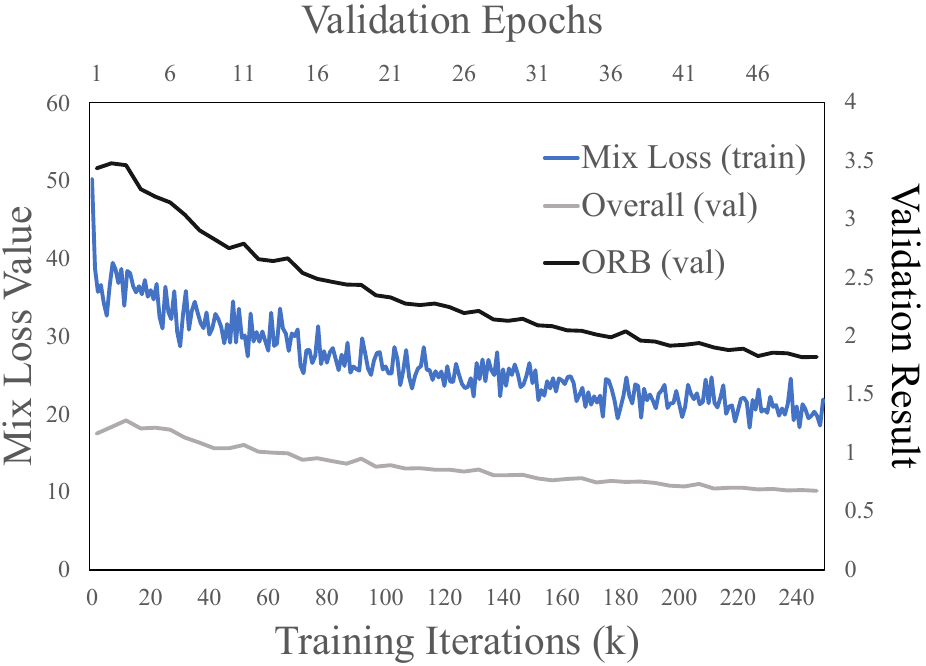}
    \caption{\textbf{Optimization procedure.} Under the supervision of the mix loss, the model exhibits efficient convergence. Due to the well-designed CPCL, the optimization effects on key points are considerably more pronounced than those on the overall frame.}
    \label{fig:loss plot}
\end{figure}

Furthermore, the CPCL possesses similar optimization properties to the conventional photometric loss. It primarily alters the manner in which we structure frame-wide supervision over points, while the computation of estimation errors for individual points remains unchanged, similar to the classic photometric loss, a convex $p$-norm objective. As a result, the mix loss function exhibits a stable and reachable global optimum, which makes it more amenable to convergence than the other data-driven methods using $\mathcal{L}_{p}$. The optimization effects under the mix loss function are illustrated in Fig.~\ref{fig:loss plot}.

As mentioned above, we leverage the proposed CPCL forcing model $M_{\theta,\eta}$ having different attention on different types of points, which has additional parameters $\alpha_i$ that vary for different points. 
The value strategy of $\alpha_i$ decides the attention level among points. 
One reasonable concern is that we not only focus the loss on key points but also on their neighbors since key points always represent local information. 
This may lead to a slight precision drop on key points, but an improvement in overall points in exchange.
Thus, it depends on the requirements of specific tasks.
For tasks that only need to track key points, to enhance the precision of key points, a large $\alpha_i$ is applied, and small $\alpha_i$ or just zero on other points.
For those in need of precise global optical flow, we set close values of $\alpha_i$ among key points' neighborhoods.
In particular, the following strategy is implemented:

\begin{equation}
    \alpha _i=\left\{ \begin{array}{l}
	\sum_{j=1}^K{\frac{1}{\sigma \sqrt{2\pi}}\exp \left( -\frac{||i-j||_{2}^{2}}{2\sigma ^2} \right)},\ condition \\
	0,\ else\\
\end{array} \right.,
\end{equation}
where $condition: ||i-j||_2\le \frac{\mu -1}{2} \ and\ I_{n1,j}\in P_k$.
This strategy defines supervision over neighborhoods of key points within a specified range of $\mu$ and generates Gaussian weights for these supervised neighborhoods with a variance $\sigma$.
It enables the model to pay attention to the neighborhoods of key points rather than solely focusing on the key points themselves.
By defining different $\mu$ and $\sigma$, we can pay different attention to different ranges of key points' neighbors.
Increasing the value of $\mu$ allows the model to learn the optical flow estimation for key points from a larger neighborhood of key points. 
On the other hand, employing a larger $\sigma$ results in smoother Gaussian weights, promoting a more uniform learning of optical flow estimation for key points within the specified neighborhood range.
The effects of tuning these parameters can be observed in Sec.~\ref{sec:ablation study}.

\begin{figure}[t]
    \centering
    \includegraphics[width=1.0\columnwidth]{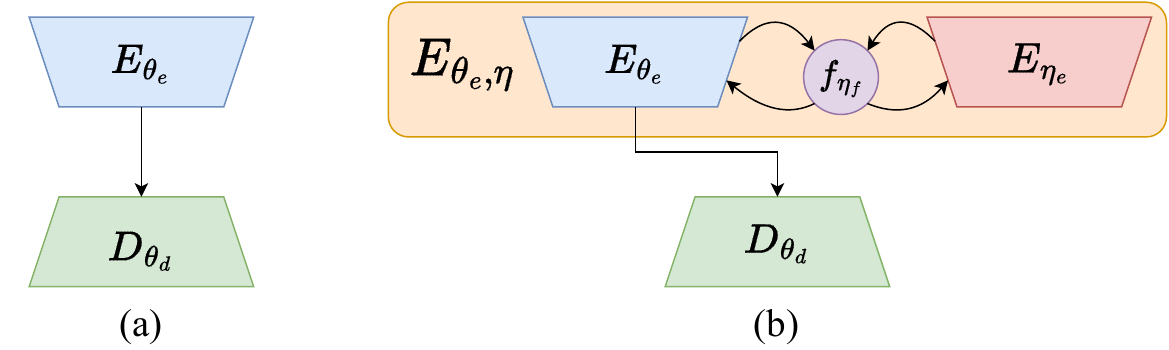}
    \caption{\textbf{Comparison between the classic architecture and the proposed conditional architecture.} (a) Classic encoder-decoder architecture. (b) The new proposed conditional architecture, by modifying the encoder $E_{\theta _e}$ into a conditioned encoder $E_{\theta _e, \eta}$ consisting of original pre-trained $\theta _e$ and condition-related $\eta$. $\eta _e$ is the encoder copy, and $\eta _f$ is the fusion module.}
    \label{fig:method}
\end{figure}

\begin{figure*}[!t]
    \centering
    \includegraphics[width=\linewidth]{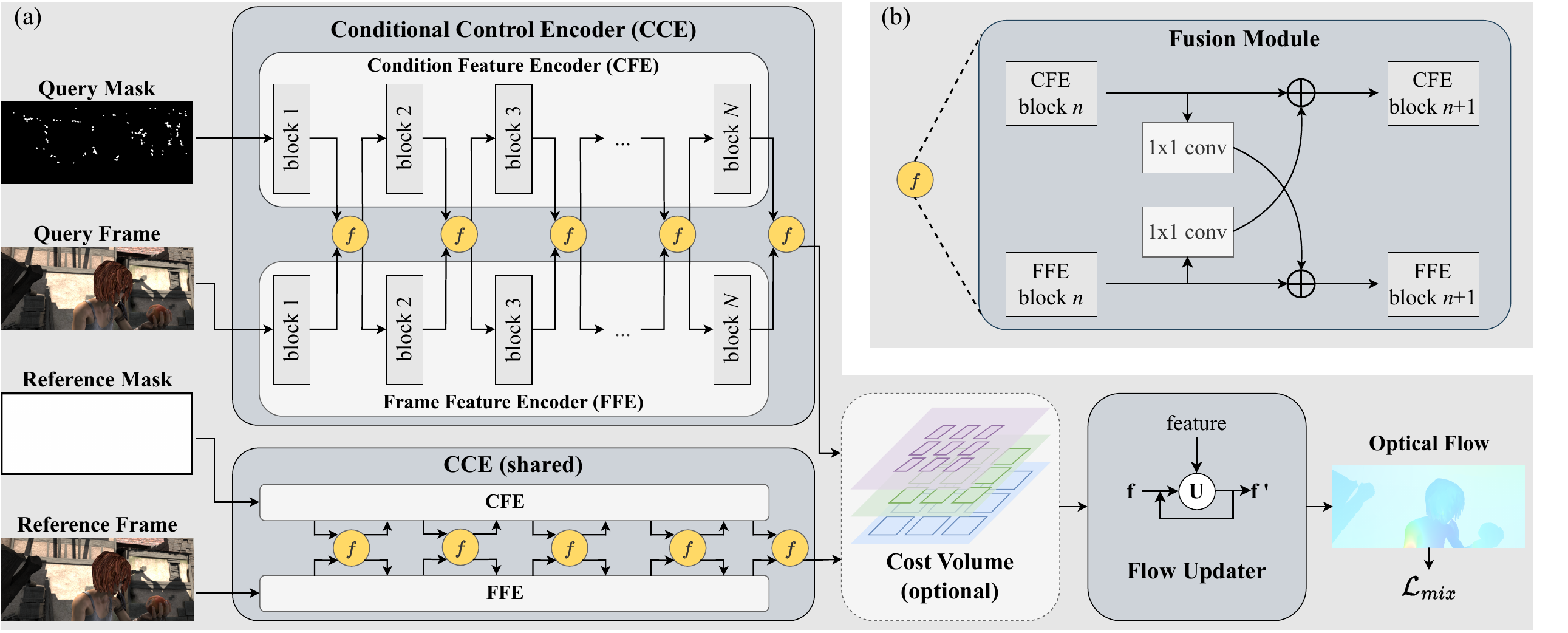}
    \caption{\textbf{Illustration of the proposed FoucsFlow framework.} (a) Architecture of FocusFlow. A Conditional Control Encoder (CCE) is proposed, which consists of a Frame Feature Encoder (FFE), a Condition Feature Encoder (CFE), and fusion modules after each stage of FFE and CFE. CFE's structure is the same as FFE but takes a conditional mask as its input. The output of CFE is used as the condition to control the estimation behavior of the whole network. Behind each stage of FFE and CFE, a fusion module is integrated which fuses the control feature and the frame feature bidirectionally and resends them into the next stage. To introduce the key point querying relationships into the network, a key point location map is set as the query mask, and a full-ones map as the reference mask. Note that the ``Cost Volume'' is optional for some approaches and not changed in the FocusFlow framework. (b) The internal structure of the fusion module. The simple and effective $1{\times}1$ convolutions are utilized as the fusion method.}
    \label{fig:network}
\end{figure*}

\subsection{Conditional Modeling}
\label{sec:conditional modeling}
In this part, a novel architecture is proposed by importing specific query relations on key points to achieve a controlling model focusing on estimating optical flow on key points, which can support most existing optical flow networks.

The new loss function $\mathcal{L}_{mix}$ proposed in Sec.~\ref{sec:loss function} requires the model to have abilities that process points diversely. 
To achieve this, conditional modeling is considered, which could convert the original model $M_\theta$ into a new model $M_{\theta,\eta}$ with a condition control $\eta$.
This model $M_{\theta,\eta}$ has original learnable parameters $\theta$ and new learnable parameters $\eta$ used for conditional modeling.

As discussed in Sec.~\ref{sec:points distribution}, we finally model $  p(\mathrm{p},\mathbf{f_{\mathrm{p}}}) = p(\mathrm{p}_k)p(\mathbf{f_{\mathrm{p}}}|\mathrm{p}_k)$, where $p(\mathrm{p}_k)$ is related to the given condition about key point's position. 
This leads to two questions, how to formalize a data-driven optical flow estimation method as a probabilistic model, and then how to develop an existing model $M_\theta$ into a new model $M_{\theta,\eta}$ which is compatible with the original structure.
For the first question, the classic data-driven models are decoupled into two parts, a feature encoder $E_{\theta _e}$ that models $p(\mathrm{p})$, and an optical flow decoder $D_{\theta _d}$, namely an optical flow updater, which models $p(\mathbf{f_{\mathrm{p}}}|\mathrm{p})$. 
This helps to answer the second question, by maintaining the original optical flow decoder $D_{\theta _d}$ and modifying the feature encoder $E_{\theta _e}$ into a conditioned feature encoder $E_{\theta _e, \eta}$ that models the $p(\mathrm{p}_k)$.

Inspired by ControlNet~\cite{zhangAddingConditionalControl2023}, which has successful usage in adding additional control to diffusion models, a similar control form for $\eta$ is utilized.
We set an encoder copy $\eta_e$ as the bypass branch to extract control information, using its output feature map in each level to control the behavior of the backbone. 
Most optical flow networks do not have powerful feature encoders like ControlNet, making the backbone locked like ControlNet and unable to sufficiently model $p(\mathrm{p}_k)$. 
In contrast, we keep it trainable and use a more sophisticated fusion module $\eta_f$ for more effective control.
This new architecture is shown in Fig.~\ref{fig:method}.

When encoding the input condition, a sparse input mask that indicates the key points' location for the query frame is implemented. 
For the reference frame, a full-one mask is employed. 
This imports an ambiguous query relation, therefore guiding the network focusing on estimating the optical flow for given queries on key points.

For training, we find it easier to learn by loading trained parameters $\theta_e$ and $\theta_d$ like ControlNet, and fine-tuning the whole model $M_{\theta, \eta}$. The reason is that training $M_{\theta, \eta}$ from start to modeling  $p(\mathrm{p}_k)$ is a hard task, but training it from a knowledgeable stage which can model $p(\mathrm{p}_o)$ is much easier. We refer the readers to Sec.~\ref{sec:ablation study} for more details.

\subsection{FocusFlow Framework}
\label{sec:framework}

In Sec.~\ref{sec:loss function} and Sec.~\ref{sec:conditional modeling}, a new loss function expression and a new architecture are presented. In this part, we implement them to construct the FocusFlow framework.
Fig.~\ref{fig:network} shows the proposed FocusFlow framework. 
We import additional query relations into the proposed framework and train the model with the proposed mix loss function. 
The binary query mask contains the key points' information, in which the positive value indicates the key points' location, while the reference mask is a full-one mask.

In the network, we propose the Condition Feature Encoder (CCE) as the encoder in the FocusFlow framework, which extracts condition information from the input mask and uses it to control the frame encoding behavior.
For the optical flow estimation method that needs an additional context encoder (\textit{e.g.}, RAFT~\cite{teed2020raft}), this is similarly supplanted by the proposed CCE.
Then the features from CFE and FFE are fused bidirectionally at each stage to enhance the control.
For the fusion modules, we apply $1{\times}1$ convolutions which is simple and effective, as the input mask and frame are well aligned.
A pair of parameter-shared CCE is applied to extract conditioned features from the query frame and reference frame separately.
These conditioned features are used for flow estimation.
During flow estimation, some original methods~\cite{teed2020raft,sun2018pwc,huang2022flowformer} use extracted features to construct a cost volume before flow updating.
For those methods, we keep this operation and use the conditioned features of CCE to build cost volume.
Lastly, the flow updater is supposed to use conditioned features or constructed cost volumes to estimate the optical flow iteratively, with no change of implementation compared to the original network.

The network is supervised using the $\mathcal{L}_{mix}$ defined in Eq.~(\ref{equ:mix loss}).
The $\mathcal{L}_{p}$ follows the form used in the original flow estimation method. 
We use key point mask to calculate related $\mathcal{L}_{cpcl}$ is calculated based on $\mathcal{L}_{p}$, and add them with hyper weight $\lambda$ into the mix loss function.
The whole network is updated by using stochastic gradient descent.

During training, the whole framework is fine-tuned by loading FFE and the flow updater with pre-trained parameters, while CFE and fusion modules are trained from scratch. 
When applying multi-stage training, it is suggested to load the whole network's parameters from the previous training stage and fine-tune them at the current training stage.

\section{Experiments}
In this section, we conduct a comprehensive variety of experiments to evaluate our proposed FocusFlow framework. 
We commence by detailing our experimental configurations in Section~\ref{sec:implementation details}. 
Subsequently, the outcomes of the four key point types on the FlyingChairs validation set are presented in Section~\ref{sec: results on various key points}, followed by the results obtained on the Sintel and KITTI validation datasets via various training stages in Section~\ref{sec:results on sintel and kitti}.
In Sec.~\ref{sec:ablation study}, several ablation studies are explored.
Lastly, our primary findings of experiments are summarized in Sec.~\ref{sec:summary}.

\subsection{Implementation Details}
\label{sec:implementation details}
\noindent\textbf{Experimental setup.}
All experiments are performed on $1{\times}$ NVIDIA A800 GPU.
For evaluation, the Average End-Point-Error (AEPE) on key points is applied which computes the mean flow error over all valid pixels. 

Since the proposed mix loss function adjusts the learning objective by changing the supervision area of CPCL and $\lambda$ in Eq.~(\ref{equ:mix loss}), we set CPCL approximate to point supervision by setting $\mu{=}1$, $\sigma{=}0.01$, and $\lambda{=}1$ for the mix loss function, for it yields the best results on key points, as will be shown in Sec.~\ref{sec:ablation study} for ablation studies. 
While the goal of this work is to boost key-point optical flow estimation on key points, it is beneficial to apply this setting.

\begin{figure}[t]
    \centering
    \includegraphics[width=1.0\columnwidth]{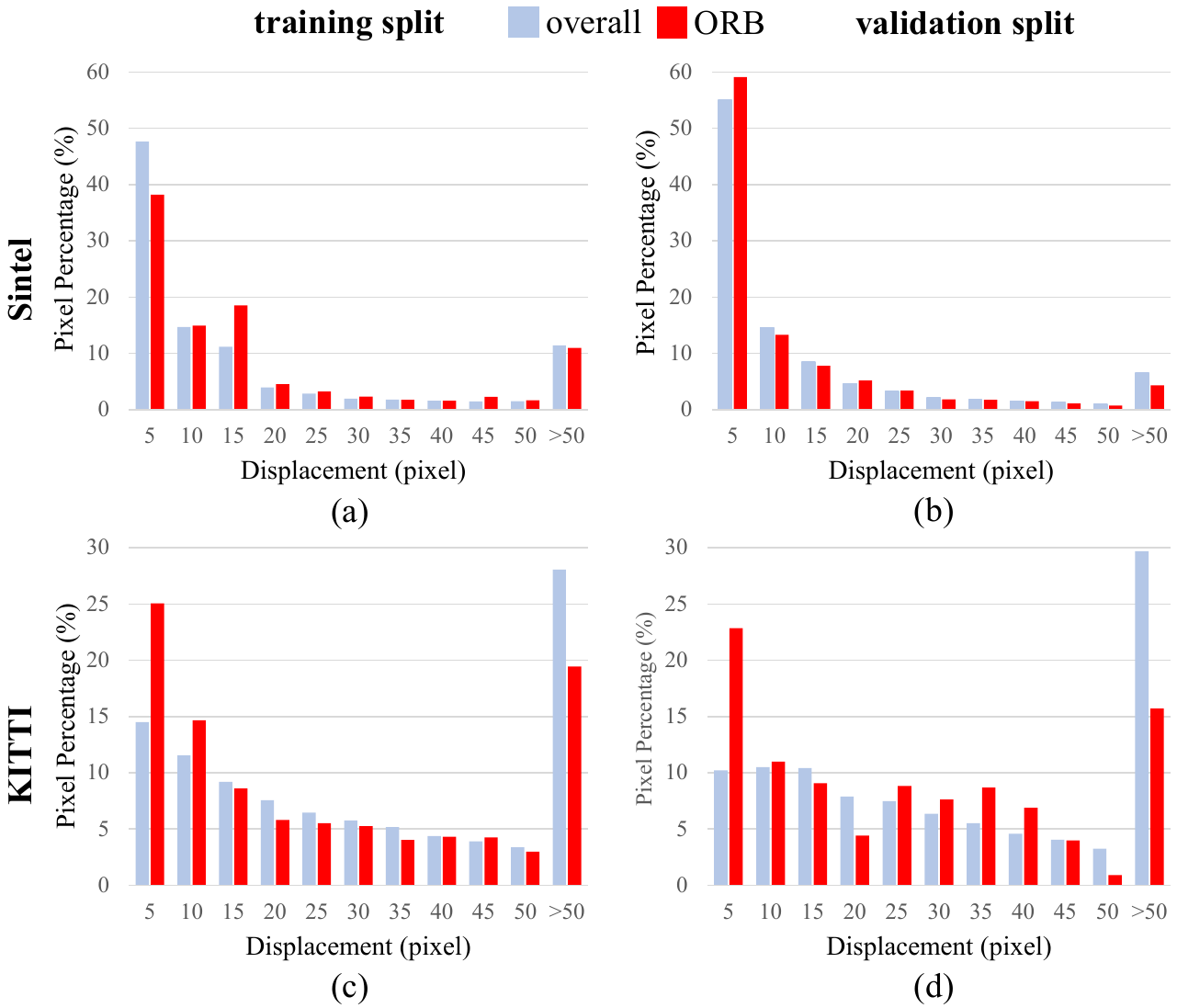}
    \caption{\textbf{Pixel displacement properties of Sintel and KITTI datasets.}
    (a) and (b) show the displacement histograms for the training split and validation split of the new Sintel dataset separately.
    (c) and (d) show the histograms for the training split and validation split of the new KITTI dataset.
    }
    \label{fig:dataset}
\end{figure}

\noindent\textbf{Datasets.}
The classical optical flow estimation datasets: FlyingChairs~\cite{dosovitskiy2015flownet}, FlyingThings3D~\cite{mayer2016large}, Sintel~\cite{butler2012naturalistic} and KITTI~\cite{geiger2013vision}, are applied in this work. 
As the ground truth of the test split of Sintel~\cite{butler2012naturalistic} and KITTI-15~\cite{geiger2013vision} is not available, to provide the accurate evaluation of AEPE on key points, we randomly split $20\%$ of the original dataset as the validation dataset. 
The displacement histogram of the split dataset is shown in Fig.~\ref{fig:dataset}.
The suffix ``-train'' denotes the split training part and ``-val'' denotes the validation part.

For training, following mainstream optical flow networks~\cite{sun2018pwc, teed2020raft, huang2022flowformer}, the model is pre-trained on the FlyingChairs dataset~(C) and then on FlyingThings~(C+T).
Then, we load the pre-trained model and fine-tune it on the mixed dataset of Sintel-train and FlyingThings~(C+T+S).
Lastly, the model is fine-tuned on the mixed dataset of Sintel-train and KITTI-train~(C+T+S+K).
In addition, the same augmentation strategy as RAFT~\cite{teed2020raft} is adopted for all models.

\noindent\textbf{Models with FocusFlow implementations.}
The FocusFlow framework has been tested on three representative optical flow networks, including PWC-Net~\cite{sun2018pwc}, RAFT~\cite{teed2020raft}, and FlowFormer~\cite{huang2022flowformer}. 
The cyclical training schedules~\cite{smith2019super} are used for the FocusFlow framework, with iterations and maximum learning rates adjusted in experiments.

For PWC-Net~\cite{sun2018pwc}, we use the PyTorch reimplementation version~\cite{pytorch-pwc} with a $7$-level pyramid.
It is trained on the FlyingChairs dataset for $1.2M$ iterations, then on C+T for $500K$ iterations, along with $S_{long}$ and $S_{fine}$ learning rate schedules introduced in~\cite{ilg2017flownet}, respectively.
While on C+T+S and C+T+S+K, the schedule proposed in~\cite{sun2019models} is applied for $300k$ iterations.
For the FocusFlow framework on PWC-Net, \textit{i.e.} FocusPWC-Net, the feature pyramid extractor is extended into CCE, where each pyramid extraction layer is treated as a single stage, followed by fusion modules.
The original warping layer, cost volume layer, optical flow estimator, and the last refiner in PWC-Net are kept the same.

For RAFT~\cite{teed2020raft}, we use its pre-trained model on FlyingChairs and FlyingThings. 
Considering that the training process on Sintel and KITTI is changed, the model is fine-tuned on C+T+S for $100k$ iterations, and then on C+T+S+K for $50k$ iterations, following~\cite{teed2020raft}. 
Due to the inclusion of an additional context encoder in RAFT, FocusRAFT is constructed by replacing both the frame feature encoder and context encoder with CCE. 
Notably, the cost volume layer and the flow updater remain unchanged.

In the case of the FlowFormer method~\cite{huang2022flowformer}, we extend it to the FocusFlowFormer approach by substituting both the frame feature encoder and the context encoder with CCE, while maintaining the integrity of other structures.
As the encoder of FlowFormer is adopted from the initial two stages of ImageNet-pre-trained Twins-SVT~\cite{chu2021twins}, the frame feature extraction knowledge prevents the CFE from learning the control from the input sparse mask. Thus, the pre-trained version is not applied to CFE. 

\begin{table}[t]
\centering
\caption{Results on various key points.}

\resizebox{\columnwidth}{!}{%
\begin{threeparttable}
\begin{tabular}{llcccc}
\toprule
Key Point Type          & Params  & ORB~\cite{rublee2011orb}  & SIFT~\cite{lowe2004distinctive} & GF\tnote{*}~\cite{shi1994good} & SiLK~\cite{gleize2023silk} \\ \midrule
PWC-Net~\cite{sun2018pwc}& 9.37M  & 5.42     & 2.78     & 3.72            & 3.92         \\
FocusPWC-Net            & 11.14M  & \textbf{5.05}     & \textbf{2.58}     & \textbf{3.59}            & \textbf{3.80}         \\ \midrule
RAFT~\cite{teed2020raft}                    & 5.25M   & 3.08 & 1.37 & 2.29        & 2.40         \\
FocusRAFT               & 7.66M   & \textbf{1.82} & \textbf{1.02} & \textbf{1.27}        & \textbf{1.47}         \\ \midrule
FlowFormer~\cite{huang2022flowformer}              & 16.16M  & 2.43 & 1.08 & 1.92        & 2.01         \\
FocusFlowFormer         & 24.80M  & \textbf{1.76} & \textbf{0.84}     & \textbf{1.25}            & \textbf{1.49}         \\ \bottomrule
\end{tabular}%

\begin{tablenotes}
 \footnotesize
 \item[*] GF represents GoodFeaturesToTrack from \cite{shi1994good} which being used as key points in VINS-Mono\cite{qin2018vins}.
 
\end{tablenotes}

\end{threeparttable}
}
\label{table:experiment_keypoint_type}
\vskip -1.6\baselineskip plus -1fil
\end{table}

\begin{table*}[t!]
\centering
\caption{Results of Overall's and ORB's EPE and their $L_c$ on Sintel and KITTI datasets.}
\resizebox{0.8\linewidth}{!}{%
\begin{threeparttable}
\begin{tabular}{cllccccccccc}
\toprule
\multirow{2}{*}{Stage}   & \multirow{2}{*}{Model}  & \multicolumn{3}{c}{Sintel-clean-val\tnote{*}} & \multicolumn{3}{c}{Sintel-final-val\tnote{*}} & \multicolumn{3}{c}{KITTI-val\tnote{*}} \\
                           \cmidrule(l){3-5} \cmidrule(l){6-8} \cmidrule(l){9-11}
                         &                        & \textcolor{gray}{Overall}      & ORB      & $L_c$  & \textcolor{gray}{Overall}      & ORB      & $L_c$  & \textcolor{gray}{Overall}    & ORB   & $L_c$   \\ \midrule
\multirow{4}{*}{C+T}     & PWC-Net~\cite{sun2018pwc}                & \textcolor{gray}{2.14}         & 3.63     & 2.18       & \textcolor{gray}{3.48}         & 5.71     & 2.47       & \textcolor{gray}{8.87}       & 9.70  & 2.66     \\
                         & FocusPWC-Net           &  \textcolor{gray}{2.11}  & \textbf{3.46}  & \textbf{1.47}       &  \textcolor{gray}{3.50}  & \textbf{5.59}   & \textbf{1.51}       &  \textcolor{gray}{6.98}    & \textbf{9.66}      & \textbf{1.56}     \\ \cmidrule(l){2-11}
                         & RAFT~\cite{teed2020raft}                   & \textcolor{gray}{1.08}         & 3.13     & 0.69   & \textcolor{gray}{1.65}         & 4.30     & 0.69  & \textcolor{gray}{7.85}       & 3.82  & 0.72     \\
                         & FocusRAFT              & \textcolor{gray}{1.09}         & \textbf{2.60}   & \textbf{0.21}       & \textcolor{gray}{2.16}         & \textbf{4.13}   & \textbf{0.18}       & \textcolor{gray}{8.24}       & 4.57  & \textbf{0.18}     \\ \midrule
\multirow{4}{*}{C+T+S}   & PWC-Net~\cite{sun2018pwc}                &  \textcolor{gray}{2.04}             & 3.63         & 1.65       &  \textcolor{gray}{2.84}             & 4.67         & 1.98       &  \textcolor{gray}{9.06}           & 9.59      & 1.99     \\
                         & FocusPWC-Net           &  \textcolor{gray}{1.99}     & \textbf{3.48}         & \textbf{0.96}       &  \textcolor{gray}{2.74}   & \textbf{4.30}         & \textbf{1.11}       &  \textcolor{gray}{7.47}           & \textbf{9.33}      & \textbf{1.13}     \\ \cmidrule(l){2-11}
                         & RAFT~\cite{teed2020raft}                   & \textcolor{gray}{1.30}         & 2.42     & 0.73       & \textcolor{gray}{1.97}         & 3.28     & 0.72       & \textcolor{gray}{3.54}       & 3.32  & 0.86     \\
                         & FocusRAFT              & \textcolor{gray}{1.40}         & \textbf{1.93}   & \textbf{0.28}       & \textcolor{gray}{2.12}         & \textbf{2.83}   & \textbf{0.27}       & \textcolor{gray}{4.02}       & 3.63      & \textbf{0.42}     \\ \midrule
\multirow{4}{*}{C+T+S+K} & PWC-Net~\cite{sun2018pwc} &  \textcolor{gray}{2.52}  & 4.03      & 1.23  &  \textcolor{gray}{3.37}       & 5.41             & 1.48         &  \textcolor{gray}{2.89}       & 11.00           & 1.16          \\
                         & FocusPWC-Net           &  \textcolor{gray}{2.67}   & 4.45         & \textbf{0.83}       &  \textcolor{gray}{3.49} & \textbf{5.17}         & \textbf{0.96}       &  \textcolor{gray}{2.67}  & \textbf{6.80}      & \textbf{0.86}     \\ \cmidrule(l){2-11}
                         & RAFT~\cite{teed2020raft} & \textcolor{gray}{1.33}         & 2.49     & 0.82       & \textcolor{gray}{2.08}         & 3.31     & 0.77       & \textcolor{gray}{1.84}       & 1.89  & 0.91     \\
                         & FocusRAFT              & \textcolor{gray}{1.43}         & \textbf{1.97}   & \textbf{0.25}       & \textcolor{gray}{2.15}         & \textbf{2.86}   & \textbf{0.19}       & \textcolor{gray}{2.29}       & \textbf{1.67}  & \textbf{0.18}     \\ 
                         \bottomrule
\end{tabular}

\begin{tablenotes}
 \footnotesize
 \item[*] We use the random-split $20\%$ sub-part of the original training dataset as the validation dataset, and the rest as the new training dataset.
 
\end{tablenotes}

\end{threeparttable}
}

\label{table:expriment_models}
\vskip -1.6\baselineskip plus -1fil
\end{table*}

\subsection{Results on Various Key Points}
\label{sec: results on various key points}

In this part, we conduct experiments on four types of key points, including ORB~\cite{rublee2011orb}, SIFT~\cite{lowe2004distinctive}, GoodFeaturesToTrack~\cite{shi1994good} and learning-based SiLK~\cite{gleize2023silk}, as shown in Table~\ref{table:experiment_keypoint_type}.
The ORB has been used in ORB-SLAM~\cite{mur2015orb}, and GoodFeaturesToTrack has been applied in VINS-Mono~\cite{qin2018vins}.
ORB and SIFT key points are extracted from the frames based on the default parameter settings of OpenCV.
For GoodFeaturesToTrack, we follow the settings of VINS-Mono~\cite{qin2018vins}, \textit{i.e.}, with the number of maximum corners of $500$, the corners' quality level of $0.01$, and the minimum possible Euclidean distance between the corners of $10$.
For SiLK, we apply a threshold of $0.2$ and minimum best key points of $500$.
The relative key points are extracted on all datasets under the above settings.

For experiments shown in Table~\ref{table:experiment_keypoint_type}, models are trained and validated on the FlyingChairs dataset.
For FocusPWC-Net, we conduct training over $1.2$ million iterations, employing a maximum learning rate of $1{\times}10^{-4}$.
On the other hand, for FocusRAFT and FocusFlowFormer, {the models are initialized with pre-trained weights from the original network trained on the FlyingChairs dataset. }
We then proceed to train both models for $250k$ iterations. However, the maximum learning rate is set at $4{\times}10^{-4}$ for FocusRAFT and $2.5{\times}10^{-4}$ for FocusFlowFormer.

For all used key points and models, the FocusFlow framework reveals better performance, with improvements of up to ${+}40.9\%$ for ORB points, ${+}22.2\%$ for SIFT points, ${+}44.5\%$ for GF points, and $38.7\%$ for SiLK points.
All maximum improvements are observed on FocusRAFT.
This is attributed to the better encoder than FocusPWC-Net and the more fusion stages than FocusFlowFormer.
On the other hand, equipped with the most powerful encoder and a transformer-like architecture, FocusFlowFormer yields the most competitive performance.
Given the trade-off between the increased number of parameters and the achieved improvement, the FocusRAFT model is considered the most favorable choice for practical applications.
These experimental results demonstrate the significant improvement of the FocusFlow in precision across various types of key points, and its promising potential which makes it well-suited to a broader range of key points with enhanced efficiency and effectiveness in the future.

\subsection{Results on Sintel and KITTI}
\label{sec:results on sintel and kitti}

In this part, the multi-stage training results are reported, by evaluating the model on Sintel-val and KITTI-val datasets. 
As the FocusFlow framework delivers an emphasis on key points, in some situations, the overall AEPE reveals a slight increase in exchange, which has been discussed in Sec.~\ref{sec:loss function}.
Considering that our goal is to boost the optical flow estimation on key points, the AEPE on ORB key points is the main result we focus on and it represents the upper limit of the FocusFlow framework.
For FocusPWC-Net, it is trained on three stages for $500k$, $300k$, and $300k$ iterations respectively, with the same maximum learning rate of $1{\times}10^{-4}$.
When employing FocusRAFT, it is trained on C+T for $250k$ iterations with maximum learning rate of $4{\times}10^{-4}$, on C+T+S for $100k$ iterations with maximum learning rate of $1.25{\times}10^{-4}$, and for $50k$ iterations with maximum learning rate of $1{\times}10^{-4}$.
Throughout each stage, the model parameters are initialized with pre-trained weights from the preceding stage.

As FlyingChairs, FlyingThings, and Sintel are all synthetic datasets, the performances on the Sintel-val set are reported when applying C+T and C+T+S.
On C+T+S+K, we focus on both Sintel-val and KITTI-val, and the results are shown in Table~\ref{table:expriment_models}.
FocusFlow framework shows its remarkable performance compared with original networks, with up to ${+}20.8\%$ improvement on the Sintel-clean-val set for FocusRAFT, ${+}13.7\%$ on the Sintel-final-val set for FocusRAFT, and ${+}38.1\%$ on the KITTI-val set for FocusPWC-Net.
For FocusRAFT, it outperforms all of the other models on all noticed datasets, with the $1.93$ of the best EPE precision of ORB on the Sintel-clean-val set, $2.83$ on the Sintel-final-val set, and $1.67$ on the KITTI-val set.
FocusPWC-Net has relatively more satisfied results compared to the PWC-Net, with the $3.46$ of the best EPE precision of ORB on the Sintel-clean-val set, $4.30$ on the Sintel-final-val set, and $6.80$ on the KITTI-val set.

\begin{figure*}[t!]
    \centering
    \includegraphics[width=\linewidth]{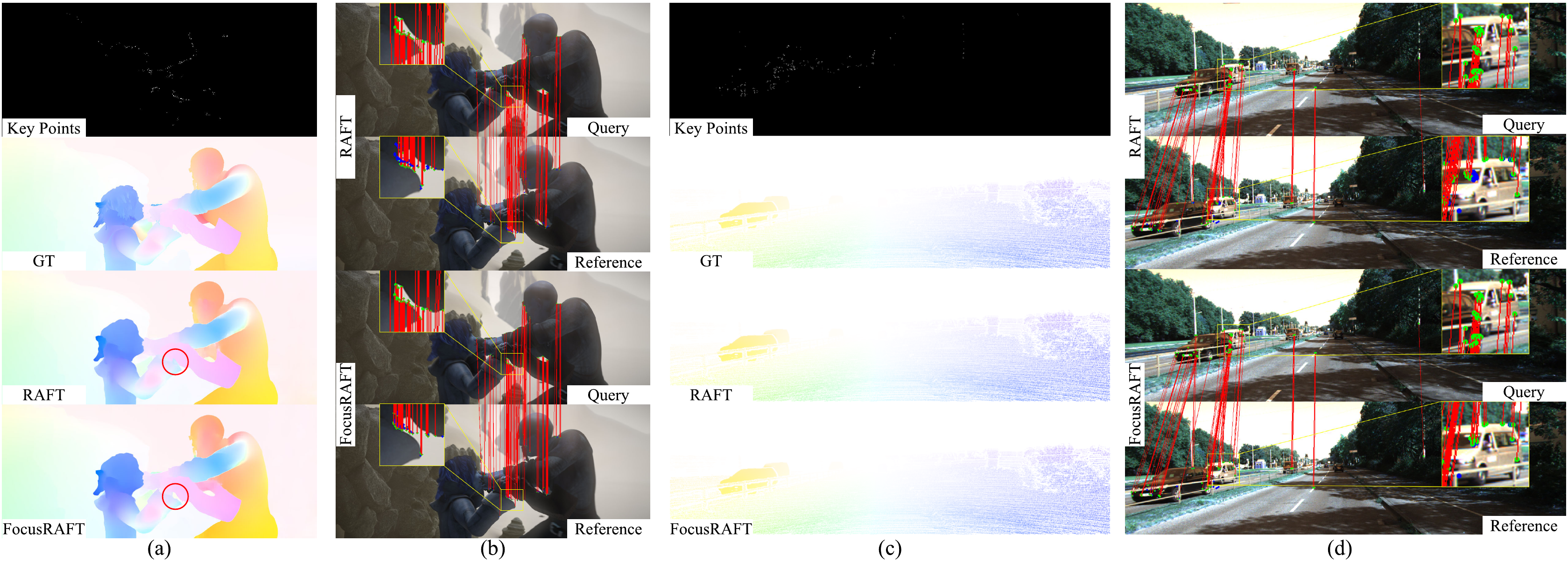}
    \caption{\textbf{Qualitative comparison of RAFT and FocusRAFT on Sintel-val and KITTI-val sets.} 
    (a) and (c) visualize optical flow estimation results of RAFT and FocusRAFT, with red circles highlighting the details.  
    (b) and (d) illustrate the key points matching results between the query frame and the reference frame.
    Green points in both the query frame and the reference frame represent the matched points, whereas blue points in the reference frame represent the ground truth.
    (a) and (b) are from the Sintel-val set, whereas (c) and (d) are from the KITTI-val set.
    }
    \label{fig:qualitative}
\end{figure*}

Additionally, {a novel metric denoted as $L_c$ is introduced}, to quantify the capacity of encoding two point sets into a unified feature space. 
This metric is computed as the Euclidean distance between the centroids of overall points and the ORB points in the embedding space of the CCE. 
Lower values of $L_c$ indicate more favorable outcomes as they reflect a superior encoder performance in prioritizing key points while also considering the entire point set.
Across all the examined datasets, the FocusFlow framework consistently achieves notably lower values of $L_c$. 
This consistent pattern underscores the advantageous role of the CCE in prioritizing the representation of key points.

Furthermore, while our primary emphasis lies in the outcomes related to key points, it is noteworthy that the performance of the FocusFlow framework in regard to overall points remains competitive with the original network, and even demonstrates superiority in certain instances. 
In cases where precision is compromised for the entire frame, the gains in precision for key points are notably significant.

As depicted in Fig.~\ref{fig:qualitative}, the FocusFlow framework yields improved optical flow estimation results, coupled with enhanced key point matching outcomes. 
This provides empirical evidence for the viability and effectiveness of the proposed methods.

\subsection{Ablation Studies}
\label{sec:ablation study}
The ablation studies primarily build upon the RAFT model~\cite{teed2020raft} and the corresponding FocusRAFT model, both evaluated using the FlyingChairs dataset. 

\begin{table}[!t]
\centering
\caption{Results of using different input key point patterns.}
\label{table:ablation_input_mask}
\resizebox{0.85\columnwidth}{!}{ 
\setlength{\tabcolsep}{20pt}
\begin{tabular}{lcc}
\toprule
Method     & Overall & ORB \\ \midrule
frame      & 0.686        & 1.966           \\
neighbor-E & 0.691        & 1.949           \\
neighbor-G & \textbf{0.671}        & 1.927           \\
context    & 0.704        & 1.897           \\
\rowcolor{gray!20}
point      & 0.688        & \textbf{1.830}           \\ \bottomrule
\end{tabular}%
}
\vskip -1.6\baselineskip plus -1fil
\end{table}

\noindent\textbf{Pattern of input condition.}
The input pattern plays a crucial role in the control effect, as it carries essential information for learning the controlling condition.
As shown in Table~\ref{table:ablation_input_mask}, we meticulously set four patterns, in which \textit{point} pattern is just binary with positive value on key points, \textit{neighbor-E} is also binary but has key points' neighbor all positive, \textit{neighbor-G} sets Gaussian weights on the key point and its neighbors, and \textit{context} means that we keep the original context information about key points' neighbors.
As ORB points incorporate $31{\times}31$ neighbor information, the neighbor range is set as a circle with a diameter of $31$.
Lastly, we set a \textit{frame} pattern, with the same input as the input frames, which has no information about the key points.
It is compared with other patterns to investigate the necessity of key points' information.

The results are reported in Table~\ref{table:ablation_input_mask}.
Notably, the \textit{point} pattern demonstrates the most favorable performance, surpassing all other patterns, including \textit{context}, \textit{neighbor-G}, and \textit{neighbor-E}. On the other hand, the \textit{frame} pattern yields the least desirable result among all the considered input patterns.
This demonstrates that the conditional control module requires precise information about the key points for better-targeted information extraction and control.
The \textit{point} pattern contains the most precise information, which is superior to the \textit{frame} with the most ambiguous description of the key points.

\begin{table}[t!]
\centering
\caption{Results of different loss function choices.}
\label{table:ablation_mixloss}

\resizebox{0.9\columnwidth}{!}{
\setlength{\tabcolsep}{10pt}
\begin{tabular}{llccccc}
\toprule
Model   & Loss Function          & Overall  & ORB \\ \midrule
\multirow{3}{*}{RAFT~\cite{teed2020raft}}    & $\mathcal{L}_p$       & 0.690         & 2.509            \\
        & $\mathcal{L}_{cpcl}$  & 0.680         & 2.199            \\
        & $\mathcal{L}_{mix}$   & 0.655         & 2.183            \\ \cmidrule(l){2-4} 
\multirow{3}{*}{FocusRAFT} & $\mathcal{L}_p$       & \underline{0.664}         & 2.415           \\
        & $\mathcal{L}_{cpcl}$  & 0.690         & \textbf{2.074}            \\
        & \cellcolor{gray!20}$\mathcal{L}_{mix}$   & \cellcolor{gray!20}\textbf{0.640}         & \cellcolor{gray!20}\underline{2.087}           \\ \bottomrule
\end{tabular}
}
\vskip -1.6\baselineskip plus -1fil
\end{table}

\noindent\textbf{Loss function choices.}
We separately fine-tune RAFT and FocusRAFT with three loss function choices, $\mathcal{L}_{p}$, $\mathcal{L}_{cpcl}$, and $\mathcal{L}_{mix}$.
Here, all models are fine-tuned for $250k$ iterations, with input masks of the $31{\times}31$ neighbor context, and fusion methods of concatenation. 
In FocusRAFT, the parameters of FFE and the flow updater are initialized by loading RAFT's weights pre-trained on the FlyingChairs dataset. 
The results under different loss functions are shown in Table~\ref{table:ablation_mixloss}.
$\lambda$, $\mu$, and $\sigma$ are set to $1$, $31$, and $5$ empirically. 
We notice that by using $\mathcal{L}_{cpcl}$ and $\mathcal{L}_{mix}$, the EPE of key points is reduced significantly for all models. 
Moreover, FocusRAFT achieves great improvements, mainly attributed to its CCE design.
In addition, the $\mathcal{L}_{mix}$ is found better than $\mathcal{L}_{cpcl}$ for RAFT, while FocusRAFT does not follow this pattern. 
This arises as the conditional control module of FocusRAFT learns exceedingly high control capabilities when employing $\mathcal{L}_{cpcl}$, which specifically concentrates on minimizing the EPE of key points, resulting in the lack of overall information extraction.
Alternatively, $\mathcal{L}_{mix}$ is a superior choice for holding a fine balance between overall points and key points. 
When transmitting from $\mathcal{L}_{cpcl}$ to $\mathcal{L}_{mix}$, significant overall accuracy improvements are observed for FocusRAFT with only a slight decrease in precision for key points.

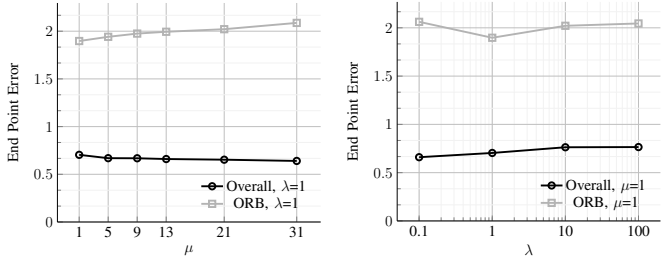
\begin{figure}[t!]
    \resizebox{0.49\linewidth}{!}{
        \begin{tikzpicture}
            \begin{axis}[
                xtick={ 1, 5, 9, 13, 21, 31}, %
                legend pos=south east,
                xticklabels={ 1, 5, 9, 13, 21, 31}, %
                xmin=-2,
                ymin=0,
                grid=both,
                grid style={line width=.1pt, draw=gray!10},
                major grid style={line width=.2pt,draw=gray!50},
                minor tick num=2,
                axis x line*=bottom,
                axis y line*=left,
                width=\linewidth,
                ylabel style= {align=center, font=\large},
                xlabel style = {font=\large, font=\large},
                ylabel={End Point Error},
                xlabel={$\mu$},
                yticklabel style = {font=\large},
                xticklabel style = {font=\large},
                legend style={cells={align=left}, font=\large, fill=none, draw=none},
            ]
            \addplot[mark=o, very thick, mark options={solid}, line width=1.5pt, mark size=2.5pt] plot coordinates {
                (1, 0.704) %
                (5, 0.669) %
                (9, 0.668) %
                (13, 0.660) %
                (21, 0.653) %
                (31, 0.640) %
            };
            \addlegendentry{Overall, $\lambda$=1}
        
            \addplot[mark=square, very thick, gray!60, mark options={solid}, line width=1.5pt, mark size=2.5pt %
                    ] plot coordinates {
                (1, 1.897) %
                (5, 1.941) %
                (9, 1.975) %
                (13, 1.994) %
                (21, 2.021) %
                (31, 2.087) %
            };
            \addlegendentry{ORB, $\lambda$=1}
        
            \end{axis}
    
\end{tikzpicture}
    } \hfill
    \resizebox{0.49\linewidth}{!}{
        \begin{tikzpicture}
            \begin{axis}[
                xtick={0.1, 1, 10, 100}, %
                legend pos=south east,
                xticklabels={0.1, 1, 10, 100}, %
                xmin=0,
                ymin=0,
                grid=both,
                xmode=log,
                grid style={line width=.1pt, draw=gray!10},
                major grid style={line width=.2pt,draw=gray!50},
                minor tick num=2,
                axis x line*=bottom,
                axis y line*=left,
                width=\linewidth,
                ylabel style= {align=center, font=\large},
                xlabel style = {font=\large},
                ylabel={End Point Error},
                xlabel={$\lambda$},
                yticklabel style = {font=\large},
                xticklabel style = {font=\large},
                legend style={cells={align=left}, font=\large, fill=none, draw=none},
            ]
        
            \addplot[mark=o, very thick, mark options={solid}, line width=1.5pt, mark size=2.5pt] plot coordinates {
                (0.1, 0.660) %
                (1, 0.704) %
                (10, 0.763) %
                (100, 0.765) %
            };
            \addlegendentry{Overall, $\mu$=1}
        
            \addplot[mark=square, very thick, gray!60, mark options={solid}, line width=1.5pt, mark size=2.5pt] plot coordinates {
                (0.1, 2.062) %
                (1, 1.897) %
                (10, 2.021) %
                (100, 2.045) %
            };
            \addlegendentry{ORB, $\mu$=1}
        \end{axis}
    
\end{tikzpicture}
    }
    \caption{\textbf{Ablation study on $\mu$ and $\lambda$}.
    We report the ablation result of different choices of $\mu$ and $\lambda$. 
    \textbf{(Left)} As $\mu$ increases, the EPE of overall points decreases slightly, while the ORB key points' EPE increases significantly.
    \textbf{(Right)} We compare with a set of $\lambda$ choices. ORB key points' EPE reaches a minimum when $\lambda{=}1$. Overall points' EPE increases continuously.
    }
    \label{fig:mixloss}
\end{figure}

We have further investigated FocusRAFT with $\lambda$ and $\mu$, as illustrated in Fig.~\ref{fig:mixloss}, by fixing $\sigma{=}(\mu{-}1)/6$, and specifically $\sigma{=}0.01$ for $\mu{=}1$ to make it approximate to the point supervision approach.
As $\mu$ increases, there is a notable decrease in the EPE for overall points, accompanied by a corresponding increase in the EPE for ORB points. This behavior is attributed to the broader range of supervision introduced by higher values of $\mu$, which brings about a closer optimization effect between $\mathcal{L}_{mix}$ and $\mathcal{L}_{p}$. 
The right part of Fig.~\ref{fig:mixloss} shows the relation between EPE and $\lambda$. 
The lowest EPE is found at $\lambda{=}1$, which offers well-balanced supervision between overall points and ORB points.
Moreover, $\lambda$ within the range of $[0.1,1]$ serves as a tuning factor for the training objective. 
This enables the model to strike a fine balance between learning optical flow estimation specifically for key points and learning estimation for the entire frame.

\noindent\textbf{Fusion methods.} 
Several classic and practical fusion methods are tested, as they are computationally simple and scalable.
In this study, all input key point patterns are set as binary point masks.
Methods \textit{conv}, \textit{concat}, \textit{SA}~\cite{woo2018cbam}, and \textit{CA}~\cite{hu2018squeeze} denote the applications of $1{\times}1$ convolutions, channel-wise concatenation, spatial attention, and channel attention separately, with bi-direction fusion implemented.
Besides, the method \textit{conv-unidirection} only involves the fusion path from CFE into FFE, which is the same as ControlNet.

Studies of these methods on the FlyChairs dataset are presented in Table~\ref{table:ablation_fusion_methods}. 
Though channel attention achieves the highest performance, it requires more than twice the additional learning parameters to achieve a marginal improvement of $0.7\%$ compared to $1{\times}1$ convolutions.
Considering both training cost and precision, the $1{\times}1$ convolutions are preferred as the fusion method.

\begin{table}[!t]
\centering
\caption{Results of different fusion methods.}
\label{table:ablation_fusion_methods}

\resizebox{0.8\columnwidth}{!}{%
\begin{tabular}{lccc}
\toprule
Method & Extra Params & Overall & ORB \\ \midrule
conv-unidirection & \textbf{2.34M}  & 0.691 & 1.910          \\
\rowcolor{gray!20}
conv   & \underline{2.41M}       & \textbf{0.676}       & \underline{1.822}           \\
concat & 2.67M       & 0.688       & 1.830           \\
SA~\cite{woo2018cbam}     & 9.33M       & 0.696       & 1.874           \\
CA~\cite{hu2018squeeze}     & 9.37M       & \underline{0.682}       & \textbf{1.809}           \\ \bottomrule
\end{tabular}%
}
\vskip -1.6\baselineskip plus -1fil
\end{table}

\noindent\textbf{Single-stage training.}
We further investigate the best single-stage training method for the FocusFlow framework.
Two training method choices are validated both for RAFT and FocusRAFT including~\textit{training from scratch} and~\textit{fine-tune}.
Additional methods~\textit{prompt-tune} for FocusRAFT freezes the original RAFT part and just trains CFE and fusion modules, and \textit{fine-tune (branch init)} means that CFE is loaded with the same pre-trained parameters as FFE. 
As shown in Table~\ref{table:training methods}, using fine-tuning does not lead to much improvement than training from scratch on RAFT, but significant improvement on FocusRAFT, showing a higher upper limit of the FocusFlow framework.
Loading FFE's pre-trained parameters as CFE's initial value is found helpless in learning CFE.
One hypothesis is that the learning goals for these two modules are not the same.
FFE is required to learn to extract features from the frame, while CFE is required to learn control from the input mask.
\textit{prompt-tune} does not yield as good results as ControlNet or other networks in that only conditional control parts are trained. 
This is partially due to the change of optimization objective and poor generalization ability of the original model since it is based on CNNs with only $5.25M$ learnable parameters.

\begin{table}[t]
\centering
\caption{Results of using different training methods.}
\label{table:training methods}

\resizebox{\columnwidth}{!}{%
\begin{tabular}{llccc}
\toprule
Model   & Method             & Params     & Overall & ORB \\ \midrule
\multirow{2}{*}{RAFT~\cite{teed2020raft}}    & train from scratch & 5.25M      & 0.756        & 2.073            \\
        & fine-tune          & 5.25M      & \textbf{0.701}        & 2.027            \\ \cmidrule(l){2-5}
\multirow{4}{*}{FocusRAFT} & train from scratch & 7.92M      & 0.741        & 2.045            \\
        & \cellcolor{gray!20}fine-tune          & \cellcolor{gray!20}7.92M      & \cellcolor{gray!20}\textbf{0.704}        & \cellcolor{gray!20}\textbf{1.897}            \\
        & fine-tune(branch init)          & 7.92M      & 0.709        & 1.981            \\
        & prompt-tune        & 3.41M      & 0.820        & 2.379            \\ \bottomrule
\end{tabular}%
}
\vskip -1.6\baselineskip plus -1fil
\end{table}

\begin{table}[h]
\centering
\caption{Studies on multi-stage training.}
\label{table:ablation_multi_stage}
\resizebox{0.8\columnwidth}{!}{

\begin{threeparttable}
\begin{tabular}{lcccc}
\toprule
\multirow{2}{*}{Method} & \multicolumn{2}{c}{Sintel-clean\tnote{*}} & \multicolumn{2}{c}{Sintel-final\tnote{*}} \\
\cmidrule(l){2-3} \cmidrule(l){4-5}
                        & Overall           & ORB          & Overall           & ORB          \\ \midrule
T(R)+N                     & 1.67             & 2.29       & 3.77             & 3.75             \\
T(R)+C(F)                     & \textbf{1.39}             & 2.13        & 3.25             & 3.51             \\
\rowcolor{gray!20}
C(F)+C(F)                     & 1.45             & \textbf{2.07}        & \textbf{3.19}             & \textbf{3.50}            \\
\bottomrule
\end{tabular}

\begin{tablenotes}
 \footnotesize
 \item[*] We perform validation on the entire Sintel training dataset.
 
\end{tablenotes}

\end{threeparttable}
}
\end{table}

\noindent\textbf{Multi-stage training.}
Optical flow estimation methods often apply multi-stage learning, by pre-training on synthetic datasets and fine-tuning on the specific realistic dataset. 
We have studied how to aggregate different knowledge from pre-training. 
The model is evaluated on the entire Sintel dataset, using FlyingChairs pre-trained parameters and training on the FlyingThings dataset.
Method \textit{T(R)+N} means the RAFT's pre-trained parameters on FlyingThings are loaded for FFE and the flow updater of FocusRAFT, and CFE is trained from scratch.
\textit{T(R)+C(F)} also means the RAFT's pre-trained parameters on FlyingThings are loaded, while CFE loads the pre-trained parameters of FocusRAFT on FlyingChairs.
\textit{C(F)+C(F)} indicates we load the whole FocusRAFT's parameters pre-trained on FlyingChairs, without the need for RAFT training on FlyingThings. 
As shown in Table~\ref{table:ablation_multi_stage}, \textit{C(F)+C(F)} reveals the best result, which means pre-training on key points helps to learn condition control.

\subsection{Summary}
\label{sec:summary}
The extensive experiments illustrate the critical points in the FocusFlow framework for boosting optical flow estimation on key points. We summarize the following primary findings of experiments:
\begin{itemize}
    \item FocusFlow framework shows great performance and scalability for most existing optical flow estimation methods and key points, with up to ${+}38.1\%$ precision improvement on ORB points on the KITTI-val dataset.
    \item A novel metric is introduced, denoted as $L_c$, to assess the capability of encoding two sets of points into a unified feature space. Remarkably, the FocusFlow framework achieves the most favorable results on this metric.
    \item The input condition using the \textit{point} pattern helps the model to learn more about controlling information extraction of FFE.
    \item Under the supervision of the proposed mix loss function, the model exhibits significant improvement in precision on key points.
    \item Through adjustment of $\mu$ and $\lambda$ in the mix loss function, the optimization direction can be readily fine-tuned to accommodate either comprehensive frame-wide estimation or specialized point-specific estimations.
    \item The $1{\times}1$ convolutions prove to be the simplest and most powerful choices for feature fusion in the FocusFlow.
    \item We verify that loading a pre-trained sub-module and then fine-tuning is the most powerful training method for single-stage training, while whole-module fine-tuning is the best solution for multi-stage training.
\end{itemize}

These findings collectively emphasize the strength and potential of the FocusFlow framework.
We believe that the FocusFlow framework can serve as an inspiration for applications in autonomous driving, SLAM, and object tracking, particularly from the perspective of key points.

\section{Conclusion}
In this paper, we propose FocusFlow, a framework that effectively enhances existing data-driven optical flow methods for estimating optical flow on key points.
Based on the consideration of treating points of the scene as a distribution of key points, a new modeling method has been put forward which requires learning a prior related to key points.
Then, CPCL is proposed for diverse point-wise supervision on the frame, to adapt the needs of different procedures for different points, and combine it with normal photometric loss function into a mix loss function.
To explicitly learn the priors of key points, a controlling model is presented that replaces the classic feature encoder with the newly proposed CCE, which consists of an FFE and a CFE.
Specifically, FFE extracts dense features from the input frames, and CFE explores controlling the feature-extracting behavior of FFE through bi-direction feature fusion after each feature extraction stage of FFE.
We use the proposed mix loss function and controlling model to construct the FocusFlow framework, with extensive experiments verifying compelling precision improvements on key points and on-par or superior accuracy on the whole frame, along with great scalability for most optical flow networks and key points.

This framework serves as a novel and generic solution to tasks that need key points' motion information, which is not strictly tied to a specific network architecture and keeps the characteristics of the original model design.
Moreover, the point-based modeling approach offers novel insights into addressing issues of varying point representations, employing point-based modeling techniques to explicitly learn the priors associated with these points.
Looking ahead, we intend to broaden the scope of our method to encompass other tasks in driving scene comprehension that demand enhanced local precision, such as monocular depth estimation.
Nevertheless, this methodology exhibits partial constraints due to its explicit dependence on key point information, as evidenced by the necessity of an input mask indicating key point locations.
This setup may introduce a minor computational overhead to detect the key points before network inference.
We anticipate and encourage further research and exploration in this area.

\bibliographystyle{IEEEtran}
\bibliography{bib}

\end{document}